# Methods to Estimate Large Language Model Confidence

Maia Kotelanski[1], Robert Gallo MD[2], Ashwin Nayak MD MS[3], Thomas Savage MD[3]

[1] Duke University, Durham, NC, USA
[2] Department of Medicine, Palo Alto Veterans Affairs Medical Center, Palo Alto, CA USA
[3] Department of Medicine, Stanford University, Stanford, CA, USA

## Abstract

**Objective**
Large Language Models (LLMs) have difficulty communicating uncertainty, and often sound confident even when incorrect, which is a significant obstacle to applying LLMs to complex medical tasks. This study evaluates methods to measure LLM confidence when suggesting a diagnosis for challenging clinical vignettes.

**Materials and Methods**
GPT-4 was asked a series of challenging case questions using Chain of Thought (CoT) and Self Consistency (SC) prompting. Multiple methods were investigated to assess model confidence and evaluated on their ability to predict the model's observed accuracy. The methods evaluated were Intrinsic Confidence, SC Agreement Frequency and CoT Response Length.

**Results**
SC Agreement Frequency correlated with observed accuracy, yielding an Area under the Receiver Operating Characteristic Curve (ROC AUC) of 0.77, while GPT-4 Intrinsic Confidence demonstrated an ROC AUC of 0.71 and CoT Length analysis an ROC AUC of 0.59.

**Discussion**
SC agreement achieves the best discrimination for model confidence. Model Intrinsic Confidence and CoT Response Length exhibit a weaker ability to differentiate between correct and incorrect answers, preventing them from being reliable and interpretable markers for model confidence.

**Conclusions**
GPT-4 has a limited ability to assess its own diagnostic accuracy. SC Agreement Frequency is the most useful method to discern GPT-4 confidence.

Key Words: Large Language Models, Artificial Intelligence, Natural Language Processing, Medical Diagnosis, Uncertainty, Confidence

## Introduction

Large Language Models (LLMs) are Artificial Intelligence (AI) tools that have been trained on large amounts of data to interpret text passages. Generative AI LLMs are a special subset of LLMs that are designed to generate free-response text by predicting the most probable response to a question or task.

Naturally there is significant interest in the application of Generative AI LLMs to medicine, with many hypothesizing LLMs could one day assist physicians with complex clinical tasks such as the diagnosis of patients or treatment selection.[1–4] Nevertheless one of the major obstacles to the use of Generative AI LLMs in patient care is their inability to communicate uncertainty when providing a response.[5–7] Generative AI LLMs are notorious for making statements that imply strong confidence in the answer even when there are obvious logical fallacies (Figure 1).[8] Accurate estimation of uncertainty is essential to engender trust when performing complex medical tasks. Therefore, the perceived inability for Generative AI LLMs to communicate answer confidence is a major obstacle for their use in clinical care.

> **Example of GPT-4 Overconfidence**
>
> A newborn infant presents with severe weakness. He was born to a G1P1 mother at 40 weeks gestation with the pregnancy attended by a midwife. The mother's past medical history is unremarkable. She took a prenatal vitamin and folic acid throughout the pregnancy. Since birth, the child has had trouble breastfeeding despite proper counseling. He also has had poor muscle tone and a weak cry. His temperature is 99.5 F (37.5 C), blood pressure is 57/38 mmHg, pulse is 150/min, respirations are 37/min, and oxygen saturation is 96% on room air. Physical exam reveals poor muscle tone. The patient's sucking reflex is weak, and an enlarged tongue is noted. An ultrasound is performed, and is notable for hypertrophy of the myocardium. What is the most likely diagnosis?
>
> **Rationale**:
> The differential diagnosis for a newborn presenting with severe weakness, poor feeding, enlarged tongue, poor muscle tone and hypertrophic cardiomyopathy includes: congenital myopathies, metabolic disorders, spinal muscular atrophy, and congenital hypothyroidism. Considering the symptomatology, congenital hypothyroidism and metabolic disorders are less likely as they typically do not cause hypertrophic cardiomyopathy. Spinal muscular atrophy is a possibility; however, hypertrophic cardiomyopathy is not typically seen in this condition. *The most likely diagnosis in this case is a congenital myopathy, more specifically, infantile botulism due to the presence of a weak cry, poor sucking reflex, generalized weakness, and hypertrophic cardiomyopathy.*
>
> **Model Answer**: Infantile Botulism  **Correct Answer**: Acid Maltase deficiency
>
> **Incorrect Logic**: Infantile Botulism is not a form of congenital myopathy and is not associated with hypertrophic cardiomyopathy.

*Figure 1. Example from Savage et al.[8] showing how GPT-4 can support a diagnosis despite incorrect logic using confident language that can mislead a user.*

## Objective

In this manuscript we experiment with strategies to assess GPT-4's confidence in answering diagnostically challenging clinical cases. Many of the described methods have been proposed in other fields of reasoning, namely arithmetic reasoning.[6,9] Here we investigate whether these methods can be applied to medicine.

## Background

The methods to estimate model confidence proposed in this investigation build upon the concepts of chain-of-thought[10] and self-consistency.[9]

Chain of thought (CoT) is a form of prompt engineering that encourages the model to break down a problem into smaller steps that can be solved one at a time. CoT has been shown to increase LLM response accuracy, facilitating more complex reasoning abilities.[10]

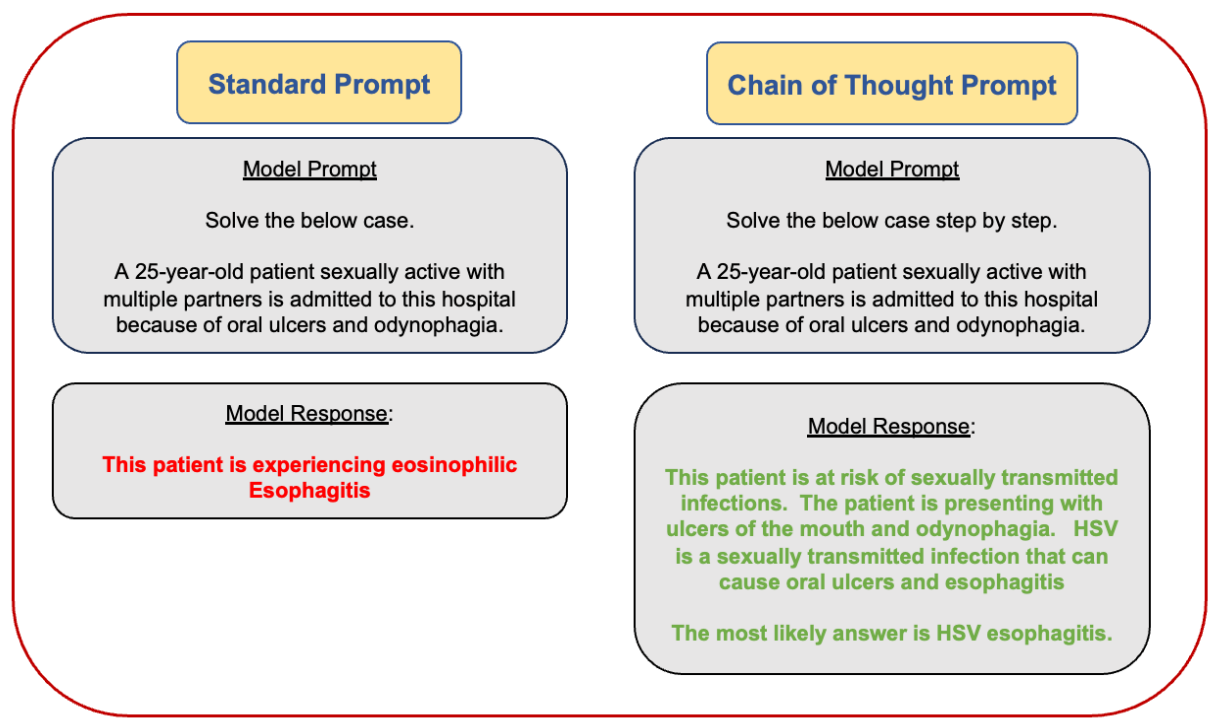

*Figure 2. Example of chain of thought prompting. The example shows an abridged prompt comparing standard prompting to chain-of-thought.*

Self-consistency is a method of running a model multiple times for the same question or task, often coupled with CoT. The model temperature parameter is set high to introduce a high degree of variability into its reasoning and produce a unique response for each run (temperature is a parameter for inducing randomness into the LLM response)..[11] The response that is then returned the most times is selected as the final answer (figure 2).

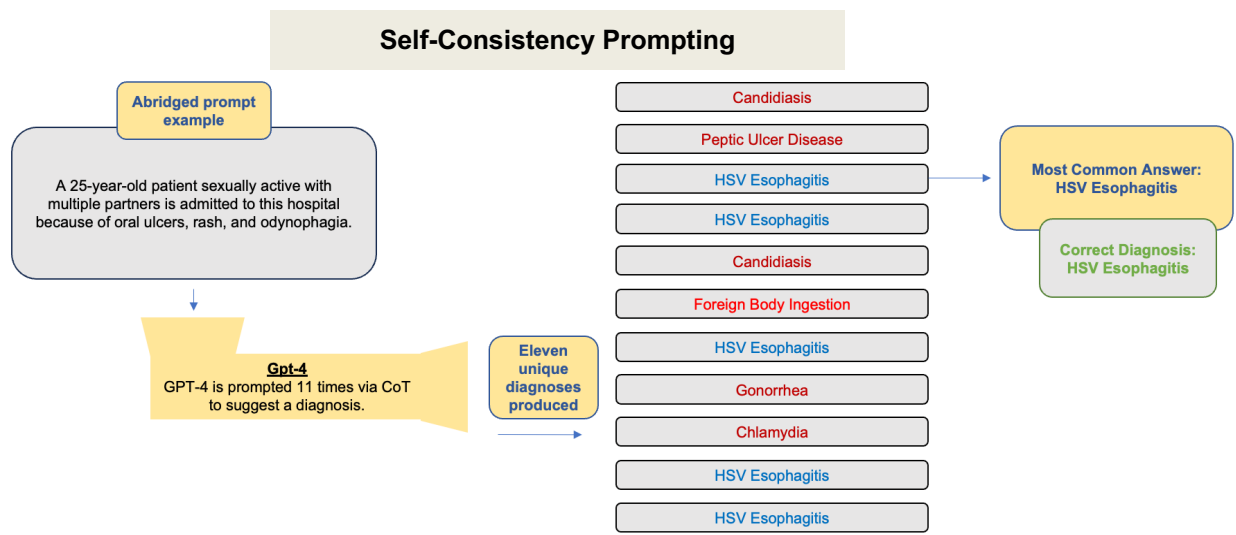

*Figure 2. Example of self-consistency prompting. The example shows an abridged prompt where the model's most common diagnosis provided the correct diagnosis in this case.*

**Confidence Assessment Methods**

This paper investigates three methods for estimating GPT-4 confidence: Intrinsic Confidence, Self-Consistency (SC) Agreement Frequency, and Chain-of-Thought Response Length. Below we conceptually describe each assessment method.

GPT-4 Intrinsic Confidence Assessment

The first method tested is Intrinsic Confidence, simply asking the model to assess its confidence in the answer that it provided. To assess Intrinsic Confidence, prompt 1 (below) was submitted to the GPT-4 Application Programming Interface (API), asking the model to rate its own confidence with a percentage from 0% to 100%, with 100% representing complete model confidence that the provided answer is the correct diagnosis.

> *"Describe what percent chance (from 0 to 100) you think the following diagnosis is the answer for the provided patient scenario provided below.*
> *Answer: {GPT-4 provided diagnosis}*
> *Case Scenario: {clinical case text}"*

*Prompt 1. This prompt was used to assess GPT-4 Intrinsic Confidence.*

Self-Consistency (SC) Agreement Frequency

The second method tested was Self-Consistency (SC) Agreement Frequency. Using self-consistency, GPT-4 was asked to run (diagnose) each case eleven times. Variation between runs was added by using a high temperature setting. The most common response was selected as the final answer, and the ratio of the number of runs producing this most common diagnosis to the total number of runs was called the SC Agreement Frequency (illustrated in Figure 3).

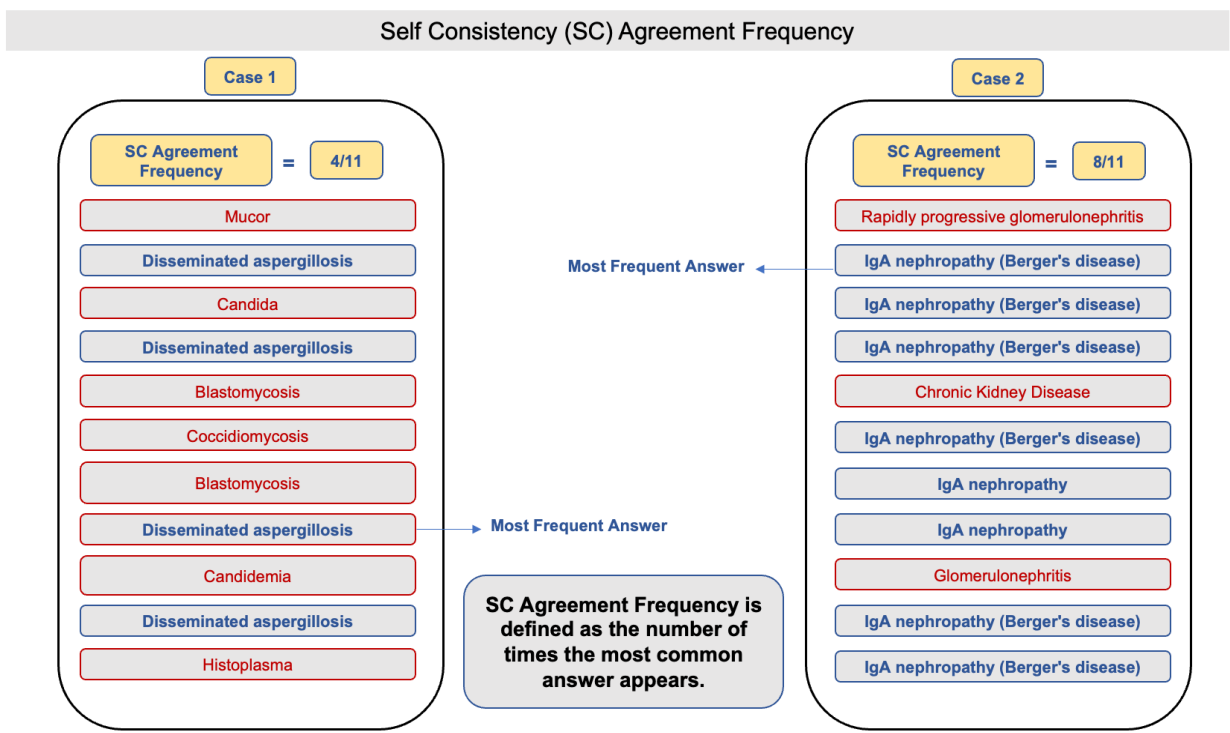

*Figure 3. Example calculation of SC Agreement Frequency shown for two cases. The figure lists out all 11 answers produced by the model and shows how the SC Frequency is the ratio of the number of times the most common answer appears to the number of answers produced.*

Chain-of-Thought (CoT) Response Length
The last method tested was CoT Response Length, which was calculated as the average length of responses in number of characters. For each case, the average response lengths were calculated for all 11 self-consistency runs. Figure 4 provides an example how CoT Response length was calculated. We hypothesize longer responses will correlate to more complex and correct answers.

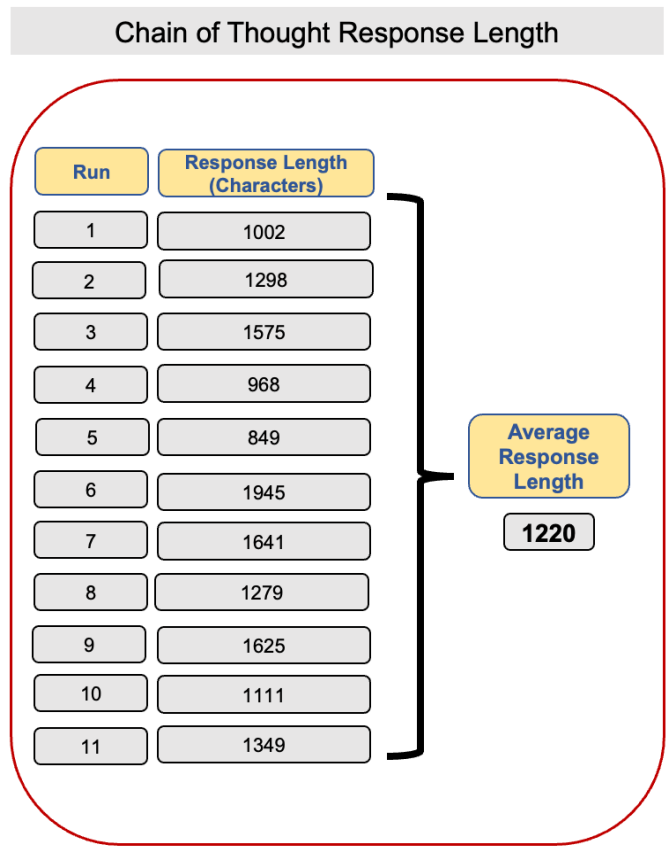

*Figure 4. CoT Response Length calculation. The response length (in number of characters) for all SC runs in each case are averaged together to get the CoT Response Length.*

**Methods**
This investigation evaluates the GPT-4 model from OpenAI.[12] GPT-4 was accessed through an API on the Google-Colab python coding platform. Full code can be found in Supplemental Information I. GPT-4 was asked a series of challenging diagnostic clinical cases and responses were recorded to evaluate strategies to estimate model confidence.

Dataset Development
The set of challenging cases on which GPT-4 was evaluated was taken from the New England Journal of Medicine (NEJM) Case Records series.[13] The NEJM Case Records series is designed as an educational resource for physicians, with each case providing a clinical case description followed by expert analysis of the case with a clinical diagnosis. We included the 191 most recently published cases in this study. A full list of all cases included (by title and DOI number) can be found in Supplemental Information II.

LLM Response Evaluation
Language model responses were judged for correctness in comparison to the provided answer from the New England Journal of Medicine by physician authors RG, AN and TS, three internal medicine attending physicians. Each question was evaluated by two blinded physicians. If there was disagreement in the grade assigned, a third evaluator determined the final grade. Any response that was felt to be equally correct and specific, as compared to the provided answer, was marked as correct. Physicians used UpToDate[14], MKSAPP[15], and StatPearls[16] to verify accuracy of answers when needed.

Chain-of-Thought Prompting Methods

The prompts used for chain-of-thought (CoT) can be found in Supplemental Information III. Prompts were the same as used by Savage et al[8] and were based on the prompts proposed by Wei et al.[10]

Self-Consistency Prompting Methods
Self-consistency was achieved by running the model (with the CoT prompt in Supplemental Information III) 11 times with a temperature setting of 1.0. Full python code is provided in Supplemental Information I. Diagnoses for each model run were manually grouped by M.K and then edited by T.S. (attending physician) into diagnoses that were considered sufficiently similar. Example groupings include combining diagnoses such as "lyme disease-induced heart block" and "lyme disease-induced carditis." The most common answer of the 11 runs was identified as the final answer. The grouping of responses was completed before authors graded answers.

Statistical Analysis
Statistical analysis was performed by calculating the Area Under the Receiver Operating Characteristic Curve (ROC AUC) for each confidence assessment method. The ROC curves were calculated using the sklearn[17] package in python.

**Results**
GPT-4's most common diagnosis by self-consistency was correct for 42% (80/191) of cases. This is a similar accuracy as compared to Kanjee et al who evaluate the same dataset.[18]

SC Agreement Frequency strongly correlated with observed accuracy (see Figure 5). In cases of complete agreement, GPT-4 was correct nearly 75% of the time and accuracy gradually declined with more observed disagreement. Cases with minimal SC agreement (2/11 or 3/11) did not diagnose any case vignettes correctly. The ROC curve for SC Agreement Frequency can be found in Figure 6, with an AUC of 0.77.

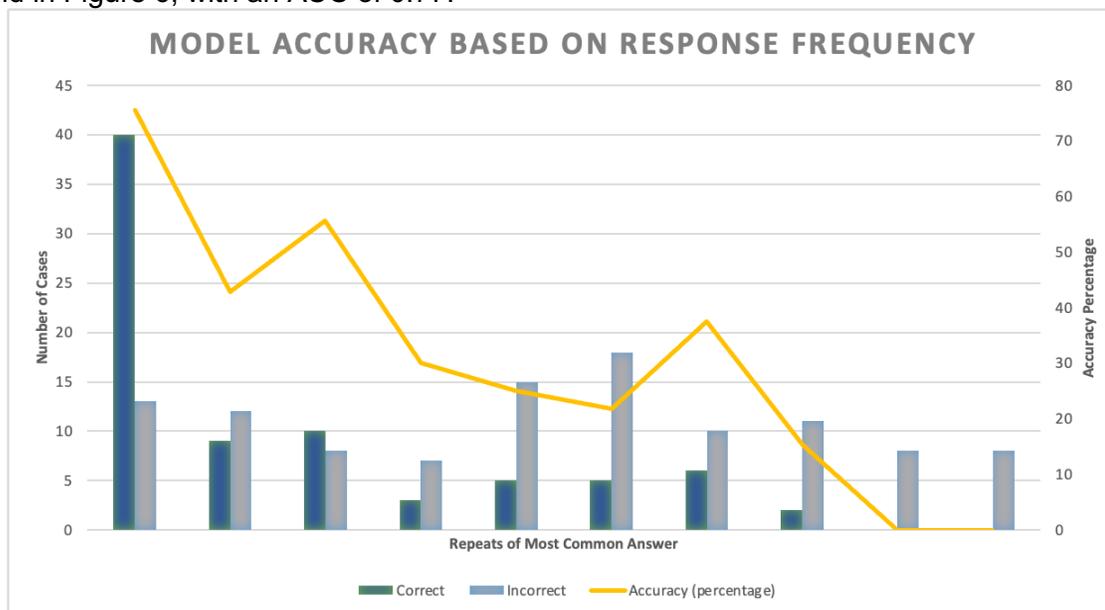

*Figure 5. SC Agreement Frequency results, demonstrating how higher degrees of SC Agreement strongly correlate with observed accuracy.*

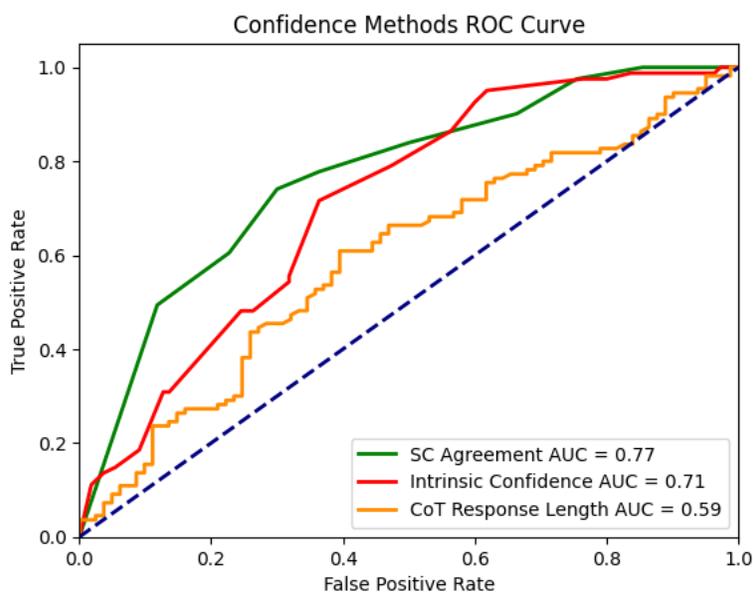

*Figure 6. Receiver Operating Characteristic Curve for SC Agreement Frequency (green), Intrinsic Confidence (red) and CoT Response Length (orange).*

When GPT-4 was asked to assess in its own Intrinsic Confidence, the model had an average predicted confidence of 69%, 79% for correct answers and 64% for incorrect answers for a net difference of 15 percentage points that is statistically significant (p value <0.001). The ROC curve for Intrinsic Confidence can be found in Figure 6 below, with an ROC AUC of 0.71.

Lastly for the CoT Response Length analysis, GPT-4 responses contained an average of 1094 characters when the model predicted the correct diagnosis and 1107 characters when predicting the incorrect diagnosis. When the model predicted an incorrect answer, its responses were on average 1.17% longer, a statistically significant difference with a p value 0.02. The ROC curve for CoT Response Length can be found in Figure 6, with an AUC of 0.59.

**Discussion**

Our investigation found SC Agreement Frequency to be the best method to assess model confidence. SC Agreement Frequency yielded the highest ROC AUC of 0.77 compared to 0.71 for Intrinsic Confidence and 0.59 for CoT Response Length. Cases where GPT-4 responses were unanimous (complete SC agreement) correlated with a high accuracy whereas lower levels of agreement progressively demonstrated less and less accuracy (Figure 5). These results agree with the preliminary hypotheses posited in other disciplines such as arithmetic.[9]

Notably SC Agreement Frequency even outperformed GPT-4's ability to self-discern answer confidence (Intrinsic Confidence). GPT-4 Intrinsic Confidence showed a statistically significant capacity to discern correct versus incorrect answers, but the degree of discernment was small. The observed 15% difference in confidence levels reported by the model between correct answers versus incorrect answers reflects the limited awareness and poor calibration LLMs have into their own uncertainty. Furthermore GPT-4 consistently over-estimated its accuracy, reporting an average confidence level of 69% percent when its true accuracy was much lower (42% observed accuracy).

CoT Response Length seemed to be the least useful metric in estimating model confidence. Decreasing length of the CoT response was statistically correlated with answer correctness, but the average difference in response length was small, only 13 characters. We interpret these results as CoT Response Length is not a sufficiently strong proxy to measure reasoning complexity, and in turn LLM confidence.

Limitations of our investigation include that only a single set of clinical vignettes (NEJM) were used and only the GPT-4 LLM by OpenAI was evaluated. Therefore, our results may not be generalizable to other test sets, medical tasks, or language models. Furthermore, our investigation consisted only of the English text and it is unknown if our results can be generalized to other non-English languages.

Overall, SC Agreement Frequency offers physicians a means to evaluate LLM response confidence and can play an integral role in how LLMs are used by physicians. LLM responses that have high or complete SC Agreement are responses that physicians can trust, while low SC Agreement should warn a physician to be skeptical of an LLM response.

**Conclusions**

The most effective method to discern GPT-4 confidence is SC Agreement Frequency. SC Agreement Frequency even outperforms GPT-4's ability to self-assess its own diagnostic confidence. SC Agreement Frequency can be a useful method for physicians who use LLMs in clinical care.

**Supplemental Information I**
Code for submitting clinical cases to the GPT-4 API with chain of thought and self-consistency prompting: "LLM_Confidence_code.py".

**Supplemental Information II**
Full list of all cases included in this study (by title and DOI number), as well as model responses and case data for each confidence method (Intrinsic Confidence, CoT Length and SC Agreement Frequency).

Excel file: "Supplemental Information II.xlsx"

**Supplemental Information III**
Below are the prompts used for the NEJM challenge cases. Each prompt includes a question-answer-rationale example based on DOI 10.1056/NEJMcpc1413303. The text for DOI 10.1056/NEJMcpc1413303 is not included in this supplemental information to respect copyright. These prompts were submitted to the GPT-4 API via the code provided in Supplemental information V.

**Traditional Chain of Thought**
Read the initial presentation of a medical case below and determine the final diagnosis. Assume that all of the relevant details from figures and tables have been explained in the text. When providing your rationale, USE STEP-BY-STEP DEDUCTION TO IDENTIFY THE CORRECT RESPONSE. After you provide your rationale, provide a single, specific diagnosis for the case in less than 10 words.

Example Case:
DOI 10.1056/NEJMcpc1413303 Text

Rationale(REMEMBER TO USE STEP BY STEP DEDUCTION):
This patient has oral ulcers, which can be associated with autoimmune and infectious diseases. The patient has a negative infectious work up and did not respond to antibiotics, which supports an autoimmune process. The patient has genital ulcers, which are associated with the autoimmune process of Behcets disease. Nodules of the legs further support an autoimmune process such as Behcets disease. Symmetric arthralgias are seen in a majority of patients with Behcets disease. Fever can be seen in systemic autoimmune processes such as Behcets disease. The rash is described as pustular, which can be seen in pathergy phenomenon, a highly specific sign for Behcets disease.
Diagnosis:
Behcets disease is the most likely diagnosis.
===

Case:
**Case Text Provided **
Rationale(REMEMBER TO USE STEP BY STEP DEDUCTION):
Diagnosis: